\documentclass[letterpaper]{article} 
\usepackage{aaai24} 
\usepackage{times}  
\usepackage{helvet}  
\usepackage{courier}  
\usepackage[hyphens]{url}  
\usepackage{graphicx} 
\usepackage{natbib}  
\usepackage{caption} 
\usepackage{algorithm}
\usepackage{algorithmic}

\usepackage{booktabs}       
\usepackage{amsfonts}       
\usepackage{nicefrac}       
\usepackage{microtype}      
\usepackage{xcolor}         

\usepackage{multirow}
\usepackage{caption}
\usepackage{subcaption}
\usepackage{graphicx}
\usepackage{amsmath}
\usepackage{amsmath,amsfonts,bm}

\usepackage{colortbl}

\newcommand{\mb}{\mathbf}

\newcommand{\mc}{\mathcal}

\usepackage{amsmath}
\usepackage{amssymb}
\usepackage{mathtools}
\usepackage{amsthm}
\usepackage{amsmath}

\DeclareMathOperator*{\argmin}{arg\,min}

\DeclareMathOperator{\E}{\mathbb{E}}

\newtheorem{theorem}{Theorem}

\theoremstyle{definition}
\newtheorem{definition}{Definition}

\theoremstyle{remark}

\usepackage{lipsum}

\usepackage{newfloat}
\usepackage{listings}
\DeclareCaptionStyle{ruled}{labelfont=normalfont,labelsep=colon,strut=off} 
\floatstyle{ruled}
\newfloat{listing}{tb}{lst}{}
\floatname{listing}{Listing}
\title{Task-Driven Causal Feature Distillation: Towards Trustworthy Risk Prediction}

\author{
Zhixuan Chu\textsuperscript{\rm 1}, Mengxuan Hu\textsuperscript{\rm 2}, Qing Cui\textsuperscript{\rm 1}, Longfei Li\textsuperscript{\rm 1}, Sheng Li\textsuperscript{\rm 2}
}
\affiliations{
    \textsuperscript{\rm 1}Ant Group\\
    \textsuperscript{\rm 2}University of Virginia\\
    \{chuzhixuan.czx,cuiqing.cq,longyao.llf\}@antgroup.com, \{qtq7su,shengli\}@virginia.edu

}

\usepackage{bibentry}

\begin{document}

\maketitle

\begin{abstract}

Since artificial intelligence has seen tremendous recent successes in many areas, it has sparked great interest in its potential for trustworthy and interpretable risk prediction. However, most models lack causal reasoning and struggle with class imbalance, leading to poor precision and recall. To address this, we propose a Task-Driven Causal Feature Distillation model (TDCFD) to transform original feature values into causal feature attributions for the specific risk prediction task. The causal feature attribution helps describe how much contribution the value of this feature can make to the risk prediction result. After the causal feature distillation, a deep neural network is applied to produce trustworthy prediction results with causal interpretability and high precision/recall. We evaluate the performance of our TDCFD method on several synthetic and real datasets, and the results demonstrate its superiority over the state-of-the-art methods regarding precision, recall, interpretability, and causality.
\end{abstract}

\section{Introduction}

The rapid development of technology not only provides a lot of convenience to people's production and life, but also brings a lot of potential risks \cite{li2022backdoor,chakraborty2018adversarial,guan2023badsam,Guan,chu2023data}, such as business risks, financial risks, medical risks, industry risks, credit risks, and so on. To prevent risks, a better way is to build an accurate risk prediction model before risks occur instead of finding a solution after the risk outbreak. Although artificial intelligence has seen tremendous recent successes in many areas \cite{luan2021review,zhu2023trustworthy,wang2023towards,shi2023aging,liu2023pharmacygpt,chen2023more}, it is often unable to produce trustworthy results on risk prediction tasks, mainly due to a lack of interpretability, no insight into cause relationships, and low precision and recall.

In fact, today’s more sophisticated deep neural network models are mostly ``black boxes'' without any knowledge of their internal workings. ``Black-box'' models are characterized by high performance but low explainability. Therefore, humans oftentimes cannot understand how machine-learned models work. Compared to general deep learning-based classification and regression tasks, the interpretability of the risk prediction model is more urgent and important. Collaborating with experts in relevant fields (e.g., finance, climate science, health care, etc.) could greatly facilitate risk prediction. In addition, risk prediction is extremely sensitive to features. For example, in the face recognition task, the result depends on the joint contribution of most features, such as noses, eyes, ears, etc. A few features cannot definitively change the recognition results, such as the face with a mask, makeup, etc. However, an abnormal fluctuation of a single feature can lead to a dramatic increase in the probability of risk occurrence. The probability of risk always hinges on a few important key features. Therefore, the expert needs to understand how the AI model works, why the model can get the output, and which feature contributes the most. Only in this way can the prediction result be trusted and adopted. 
\begin{figure*}
  \centering
  \includegraphics[width=0.95\linewidth]{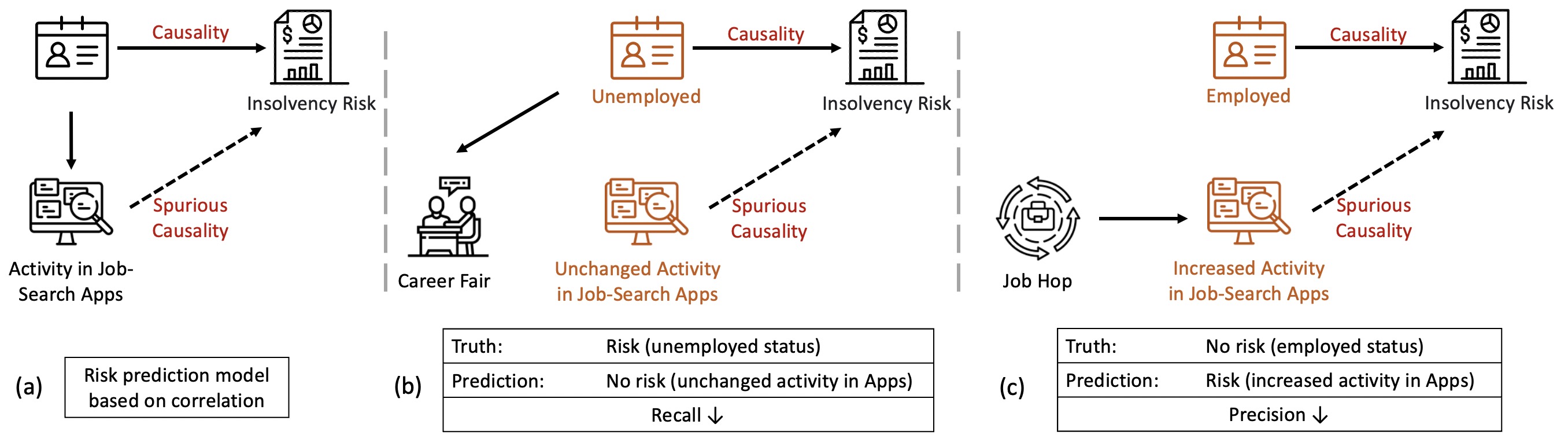}
  
  \caption{Examples of the recall and precision decrease.}
  
  \label{fig:correlation}
\end{figure*}

So far, most explainable AI models are based on correlation rather than causality. As shown in Fig. \ref{fig:correlation}, let us consider the case where we aim to utilize three variables, i.e., (1) employment status such as unemployed or employed, (2) activity in job-hunting apps such as Facebook Jobs, LinkedIn Job Search, Glassdoor, and so on, (3) gender, to predict the risk of personal insolvency. It is not hard to know by common sense that unemployed employment status can be the real cause of an increase in personal insolvency risk among these three predictors. Gender is also not directly related to the personal insolvency risk. In addition, we also know that the unemployed job status is more likely to increase the activity in job-hunting apps. Therefore, we can observe a correlation rather than a causal relationship between the risk of personal insolvency and the activity in job-hunting apps. Based on this dataset, if we run a general prediction model, it is not difficult to observe this result that the employment status and the activity in job-hunting apps are relatively important features for the risk of personal insolvency due to the spurious correlation between the true cause ``employment status'' and the fake cause ``activity in job-hunting apps''. A post-hoc explanation of this model ``correctly'' indicates that the most important features of the model are the employment status and the activity in job-hunting apps. This explanation may be deemed incorrect if we compare it against the ground-truth explanation of the underlying data. Therefore, this ``correct'' explanation based on correlation is not trustworthy. Current correlation-based approaches to explainable risk prediction models are falling short. 
The model that can figure out the causal effect may unlock trustworthy explainability.

Besides, the performance of general deep learning models is not satisfactory with respect to precision and recall. For the risk prediction task, the target label is always extremely imbalanced with skewed class proportions. Classes make up a large proportion of the negative samples (no risks) and a smaller proportion of positive samples (risks) because, in reality, the risk is always in the minority. In this circumstance, high accuracy is easily achieved by predicting every sample to be negative, even though this results in a false prediction of all positive samples. Hence, accuracy is not the be-all and end-all model metric. The expert tends to focus more on whether the model can maximumly capture the actual positive samples (actual risks) with minimum false positive samples (actual no risks). Therefore, precision and recall are key metrics for risk prediction tasks. Precision talks about how precise the model is out of those predicted positives. Recall actually calculates the proportion of the actual positives captured by the model. Therefore, in the example above, general deep models without causal explanations are not effective in ensuring high precision and recall. For example, if one person who intends to hop from job to job spends more time on job-hunting apps, the general model may identify this person as at risk of insolvency, which will decrease the precision. If an unemployed person does not seek jobs through job-hunting apps but attends several local career fairs, the model may not accurately identify him as risky due to the fact that the ``important'' feature activity in job-hunting apps is unchanged.

Without figuring out the true causal features, it is very challenging to produce trustworthy predictions with high recall and precision in the risk prediction task. Therefore, based on the particularities of risk data, we propose a Task-Driven Causal Feature Distillation model (TDCFD) to deliver trustworthy risk predictions. To our best knowledge, TDCFD is the first to incorporate the Potential Outcome Framework (POF) \cite{rubin1974estimating} into the model explanation. By utilizing the POF, the task-driven causal feature attributions are distilled from the original feature values, which represent how much contribution each feature makes to this specific risk prediction task. Trustworthy risk predictions with causal interpretability and high precision and recall can be obtained by training on distilled data. We evaluate our TDCFD method on several synthetic and real datasets, and the results demonstrate its superiority over the state-of-the-art methods in terms of precision, recall, interpretability, and causality.

\section{Preliminary}
\label{PS}

In our work, we aim to utilize causal inference to construct causal feature distillation for each feature variable. We successively treat each feature as a causal intervention (also named exposure or treatment in epidemiology) and other features as background variables to explore the causal relationship between the causal intervention and the outcome. Because the causal intervention assignment (the feature value) is not randomly distributed, the potential confounders (associated with both causal interventions and outcomes) in the background variables (other features) can lead to selection bias \cite{chu2023causal,yao2021survey,li2023machine,chu2021graph}. Thus, we need to properly adjust for background variables (other features) in order to estimate the causal effect between the causal intervention (distilled feature) and the outcome. A common approach for confounding adjustment is using the propensity score, i.e., the probability of a unit being assigned to a particular level of intervention, given the background covariates \cite{rosenbaum1983central}. In confounding adjustment, although including all confounders is important, this does not mean that including more variables is better~\cite{chu2020matching,greenland2008invited,schisterman2009overadjustment}. For example, conditioning on \textit{instrumental} variables that are associated with the intervention assignment but not with the outcome except through the intervention can increase both bias and variance of estimated causal effects~\cite{myers2011effects_instrumental}. Conditioning on \textit{adjustment} variables that are predictive of outcomes but not associated with intervention assignment is unnecessary to remove bias while reducing variance in estimated causal effects~\cite{sauer2013review}. Therefore, the inclusion of instrumental variables can inflate standard errors without improving bias, while the inclusion of adjustment variables can improve precision \cite{shortreed2017outcome,wilson2014confounder,lin2015regularization}.

\section{Methodology}

There exist several key bottlenecks in accepting deep learning models in risk prediction, such as a lack of interpretability, no insight into cause relationships, and low precision and recall. Therefore, aiming at the risk prediction task, we propose a task-driven causal feature distillation model to transform original feature values with multifaceted information into causal feature attributions for the specific risk prediction task, which describes how much contribution the value of this feature can make to the risk prediction result. After the causal feature distillation, a deep neural network is applied to predict the risk by learning feature interaction. The framework of TDCFD contains two major components: causal feature distillation and risk prediction based on causal feature attribution.

\subsection{Causal Feature Distillation}
\label{black_box}

In the causal feature distillation part, there are three major steps: relational graph construction, propensity score estimation by adaptive group Lasso, and causal feature attribution estimation. To estimate the causal effect for each feature with observational data, we successively treat each feature as a causal intervention and other features as background variables to construct a relational graph that can help to estimate the propensity score by the inclusion of covariates (confounders and adjustment variables) predictive of the outcome and exclusion of covariates (instrumental and spurious variables) unrelated to the outcome, according to the discussion in the Preliminary section.

\subsubsection{Relational Graph Construction}
\label{prediction}

Here, we need to figure out the covariates that are predictive of the outcomes while removing the covariates independent of the outcome. Consider the application of a deep neural network predicting a risk $Y$ as a function of a $d_0\times1$ vector of covariates $X \in \mc{X}$. A deep neural network is comprised of $L+1$ layers of interconnected nodes including an input layer, an output layer, and $L-1$ hidden layers. Taking $H_0=X$ for notational convenience, the $k$-th hidden layer $H_{k}=\phi_k(B_k \cdot H_{k-1} + A_k)$ is comprised of $d_k$ nodes; $k=1,\cdots,L-1$, where $\phi_k$ is an analytic activation function. The matrices $B_k \in \mb{R}^{d_{k}\times d_{k-1} }$ are comprised of unknown weights, and the $d_k \times 1$ vector $A_k$ may be regarded as a vector of intercepts. The last layer is the output layer, i.e., $H_L=\phi_L(B_L \cdot H_{L-1} + A_L)$, where the activation function $\phi_L$ depends on the type of outcome variable. Putting it all together, the output layer $f(X;\beta)=H_L$ is a composition, where $\beta$ is the collection of all neural network parameters, i.e., $B_k$ and $A_k$; $k=1,\cdots, L$. Because the deep neural network only interacts with the original covariates through the first hidden layer, the columns $\beta_1^{(1)}, \cdots,\beta_{d_0}^{(1)}$ of $B_1$  in the first hidden layer are of particular interest, as they are comprised of vectors of parameters associated with the corresponding input variables $X_1,\cdots,X_{d_0}$. Thus, the Euclidean norm $||\beta_j^{(1)} ||$ is regarded as a measure of the impact of $X_j$ on the risk outcome $Y$, where $j=1, \cdots , d_0$. 

For a general deep neural network to predict the risk, it is expressed as a nonlinear mapping $f: X \rightarrow Y$ from observed covariates $X$ to the risk outcome $Y$. In this step, we do not expect to predict the risks $Y$ very well and only aim to dig out the relationship between covariates and the risk outcome. Because the model only interacts with the original covariates through the first hidden layer, we impose a group Lasso penalty in the first layer to help select covariates predictive of the outcome, while removing covariates independent of the outcome. Here, the parameters $\beta_j^{(1)}$ connected to each input covariate $X_j$ are penalized as a group through the Euclidean norm $||\beta_j^{(1)}||$; $j=1,\cdots,d_0$, so as to simultaneously perform covariate selection. Let $\hat{y}_i=f(x_i;\beta)$ denote the predicted observed outcome of unit $i$ given input feature $x_i$. The estimator for outcome prediction with group Lasso is thus defined as:
\begin{equation}
    \widehat{\beta}_n = \argmin_{\beta} \{ R_n(\beta)+\lambda_n q(\beta)\},
    \label{Eqn: outcome}
\end{equation}
where the empirical risk function is $R_n(\beta)=\frac{1}{n}\sum_{i=1}^{n}\ell(Y_i,f(X_i;\beta))$, $\ell(Y_i,f(X_i;\beta))$ denotes the log probability density (mass) function of $Y_i$ given $f(X_i;\beta)$, $q(\beta)=\sum_{k=1}^{d_0}||\beta_j^{(1)}||$ is a penalty function, and $\lambda_n > 0$ is the tuning parameter.

Then, we aim to utilize the Potential Outcome Framework (POF) \cite{rubin1974estimating,chu2020matching,chu2022learning} to estimate the causal effect of the feature on the outcome one by one. For example, we can specify one feature $X_j$ as the intervention variable and other features $X_{-j} = X_1,..., X_{j-1}, X_{j+1},..., X_{d_0}$ as other roles (such as confounders $X_\mathcal{C}$, adjustment $X_\mathcal{P}$, instrumental $X_\mathcal{I}$, or spurious variables $X_\mathcal{S}$) under the relationship between intervention $X_j$ and risk outcome $Y$. Features $X_\mathcal{C}$ referred to as confounders influence both the intervention feature $X_j$ and the risk outcome $Y$. Features $X_\mathcal{P}$ referred to as adjustment variables are only predictive of the risk outcome $Y$. Features $X_\mathcal{I}$ referred to as instrumental variables are only predictive of the intervention feature $X_j$. Features $X_\mathcal{P}$ referred to as spurious variables are unrelated to both intervention feature $X_j$ and risk outcome $Y$. So far, according to the predictive ability of risk outcome, i.e., the group Lasso weights $||\beta_j^{(1)}||$; $j=1,\cdots,d_0$, larger penalties are automatically assigned to $X_\mathcal{I}$ and $X_\mathcal{S}$ while smaller penalties are assigned to $X_\mathcal{P}$ and $X_\mathcal{C}$. We have identified a relational graph for the feature $X_j$ (not a causal graph \footnote{Relational graph is enough to estimate a propensity score and infer the causal effect \cite{shortreed2017outcome}. Learning an accurate causal graph is a much tougher task.}).

\subsubsection{PS Estimation by Adaptive Group Lasso}

A propensity score is the probability of a unit being assigned to a particular intervention given a set of observed covariates. Propensity scores are used to reduce selection bias by equating groups based on these covariates. The propensity score is defined as the conditional probability of the intervention variable given other background variables, i.e., $e(x_j, x_{-j}) \stackrel{\text{def}}{=} P(X_j=x_j\mid X_{-j}=x_{-j})$. 

Based on the established relational graph for the intervention feature $X_j$, we adopt a deep neural network with adaptive group Lasso to estimate the propensity score, which is expressed as a non-linear mapping $g: X_{-j} \rightarrow X_j$. As discussed in Section~\ref{PS}, a propensity score estimation model should include confounders $X_\mathcal{C}$ and adjustment variables $X_\mathcal{P}$, and at the same time eliminate instrumental variables $X_\mathcal{I}$ and spurious variables $X_\mathcal{S}$. The regular Lasso forces the coefficients to be equally penalized in the $\ell_1$ penalty, regardless of the types of covariates \cite{zou2006adaptive}, and thus it cannot achieve our objective. To design a penalty function with different regularization strengths according to different types of covariates, we apply the adaptive group Lasso in outcome prediction (Eq.~\eqref{Eqn: outcome}) as the initial estimator into the propensity score estimation.

The parameters in the first hidden layer of the propensity score estimation model that directly interact with intervention feature $X_{-j}$ can also be virtually decomposed into four subsets, i.e., $[\alpha_\mathcal{C},\alpha_\mathcal{P},\alpha_\mathcal{I},\alpha_\mathcal{S}]$, where  $\alpha_\mathcal{C}  \in \mb{R}^{d_1 \times n_\mathcal{C}}$, $\alpha_\mathcal{P}\in \mb{R}^{d_1 \times n_\mathcal{P}}$, $\alpha_\mathcal{I}\in \mb{R}^{d_1 \times n_\mathcal{I}}$, and $\alpha_\mathcal{S}\in \mb{R}^{d_1 \times n_\mathcal{S}}$. The function $g$ maps the other features $X_{i,-j}$ to the explained intervention $X_{i,j}$ by a deep neural network. Here, we assume $e(x_{i,j}, x_{i,-j};\alpha)$ is the predicted propensity score, where $\alpha$ is the collection of parameters in the first layer, i.e., the probability of the feature $X_j$ of unit $i$ taking the observed value. we define the estimator of the propensity score model with adaptive group Lasso by:
\begin{equation} 
\small
\hat{\alpha}_n = \argmin_{\alpha} \bigg\{K_n(\alpha) + \theta_n q(\alpha)\bigg\},\
\label{ps estimator}
\end{equation}
where the empirical risk function is $K_n(\alpha) = \frac{1}{n}\sum_{i=1}^{n} \ell(X_{i,j},p(X_{i,-j};\alpha))$,
\begin{align}
\small
q(\alpha) & = \underbrace{\sum_{c=1}^{n_\mathcal{C}}\frac{||\alpha_{c(\mathcal{C})}||}{||\widehat{\beta}_{c(\mathcal{C})}||^\gamma}+\sum_{p=1}^{n_\mathcal{P}}\frac{||\alpha_{p(\mathcal{P})}||}{||\widehat{\beta}_{p(\mathcal{P})}||^\gamma}}_{{||\widehat{\beta}_{c(\mathcal{C})}||^{-\gamma}} \, \text{and} {||\widehat{\beta}_{p(\mathcal{P})}||^{-\gamma}} \, \text{bounded}, X_\mathcal{C} \, \text{and} \, X_\mathcal{P} \, \text{are included}} \\
& \underbrace{+\sum_{i=1}^{n_\mathcal{I}}\frac{||\alpha_{i(\mathcal{I})}||}{||\widehat{\beta}_{i(\mathcal{I})}||^\gamma}+\sum_{s=1}^{n_\mathcal{S}}\frac{||\alpha_{s(\mathcal{S})}||}{||\widehat{\beta}_{s(\mathcal{S})}||^\gamma}}_{{||\widehat{\beta}_{i(\mathcal{I})}||^{-\gamma}} \, \text{and} \,{||\widehat{\beta}_{s(\mathcal{S})}||^{-\gamma}} \, \text{inflated to infinity},X_\mathcal{I} \, \text{and} \, X_\mathcal{S} \, \text{are removed}},
\label{Eqn:ps}
\end{align}
and the tuning parameter $\theta_n > 0$ controls the trade-off between the intervention prediction and adaptive group Lasso. The power $\gamma$ is positive. 

More specifically, the feature variable $X_j$ that needs to be distilled can be binary, multiple, or continuous, so we adopt different DNN models to estimate the propensity score. For continuous variables, a mixture density network (MDN) \cite{bishop1994mixture} is adopted to model a conditional probability distribution $p(x_j|x_{-j})$ as a mixture of distributions, built within the general framework of neural networks and probability theory for working on supervised learning problems in which the target variable cannot be easily approximated by a single standard probability distribution. For binary and multiple variables, $p(x_j|x_{-j})$ is directly available. For the covariates removed in the outcome prediction with group Lasso (Eq.~\eqref{Eqn: outcome}), $\widehat{\beta}=0$. Therefore, we assume that $0/0=1$ and the corresponding $\beta$ will still converge to zero. According to the decomposition of $X_{-j}$ into $[X_\mathcal{C}, X_\mathcal{P}, X_\mathcal{I}, X_\mathcal{S}]$, the adaptive group Lasso uses the corresponding $\widehat{\beta}_{c(\mathcal{C})}$, $\widehat{\beta}_{p(\mathcal{P})}$, $\widehat{\beta}_{i(\mathcal{I})}$, and $\widehat{\beta}_{s(\mathcal{S})}$ to assign different initial weights to covariates $X_{-j}$ based on their importance in predicting outcome variable $Y$. 

In the outcome prediction model with group Lasso (Eq.~\eqref{Eqn: outcome}), the coefficients of confounders $\widehat{\beta}_{c(\mathcal{C})}$ and adjustment variables $\widehat{\beta}_{p(\mathcal{P})}$ that are predictive of the outcome should be larger than those of instrumental $\widehat{\beta}_{i(\mathcal{I})}$ and spurious variables $\widehat{\beta}_{s(\mathcal{S})}$ that are not related to the outcome. Thus, the weights ${||\widehat{\beta}_{i(\mathcal{I})}||^{-\gamma}}$ and ${||\widehat{\beta}_{s(\mathcal{S})}||^{-\gamma}}$ for instrumental and spurious variables are inflated to infinity while the weights ${||\widehat{\beta}_{c(\mathcal{C})}||^{-\gamma}}$ and ${||\widehat{\beta}_{p(\mathcal{P})}||^{-\gamma}}$ for confounders and adjustment variables are bounded. Therefore, confounders and adjustment variables can be automatically selected, and instrumental and spurious variables can be automatically removed. 

It is worth noting that we cannot distinguish confounders from adjustment variables and instrumental variables from spurious variables. In fact, there is no need to distinguish them. The estimation of propensity score benefits from the combination of confounders and adjustment variables and suffers from instrumental and spurious variables.

\subsubsection{Causal Feature Attribution Estimation}

In order to accomplish task-driven causal feature distillation, we need the response function for each feature. For each unit $i$, we postulate the existence of a set of potential outcomes, $Y_i(x_j)$, for $x_j \in X_j$, referred to as the unit-level response function. $X_j$ can be binary or multiple, and we also allow $X_j$ to be a continuous interval $[{low}^{x_j}, {high}^{x_j}]$. We are interested in the average response function, $\mu(x_j) = \E[Y_i(x_{i,j})]$. To simplify the notation, we will drop the $i$ subscript in the sequel.

In this section, we show that PS can be used to eliminate any bias associated with differences in the covariates $X_{-j}$. The approach consists of two steps. First, we estimate the conditional expectation of the outcome as a function of two scalar variables, the intervention value of $X_j$ and the PS $E$, $\sigma(x_j, e(x_{j}, x_{-j})) = \E[Y|X_j = x_j, E = e(x_{j}, x_{-j})]$. However, empirically, a one-dimensional propensity score space will lose most of the information in the data, so learning a low-dimensional propensity vector is a feasible solution \cite{chu2020matching}, which is the last layer of the propensity score estimation model. Second, we estimate the response function at a particular value of the intervention $X_j$. We average this conditional expectation over the score evaluated at the intervention level of interest ($e(x_j, X_{-j})$ rather than $e(X_j, X_{-j})$), i.e., $\mu(x_j) = \E[\sigma(x_j, e(x_{j}, X_{-j}))]$. In the binary intervention case, $x_j ={0, 1}$, and in the multiple intervention case, $x_j$ has multiple values. For continuous intervention, $x_j$ is in an interval $[low^{x_j}, high^{x_j}]$. 

In the following, we will give several definitions based on the average response function, $\mu(x_j) = \E[Y(x_j)]$ to help to explain the contribution of each feature for the specific task.

\theoremstyle{definition}
\begin{definition}({Causal Interventional Expectation}).
In the potential outcome framework (POF) \cite{rosenbaum1983central}, the Causal Interventional Expectation $\E[Y(x_j)]$ is defined as the expectation of all potential outcomes overall populations given the specific value $x_j$ of feature $X_j$.
\label{def: causal importance}
\end{definition}

This is similar to the $\E[Y|do(X_j=x_j)]$ defined in the structural causal model. $do(\cdot)$ operator simulates physical interventions by fixing $X_j$ with a constant $x_j$, while keeping the rest of the features unchanged. A causal response curve, a figure illustrating the expectation of outcome across a specific feature, can be derived from causal interventional expectation.

\theoremstyle{definition}
\begin{definition}({Causal Feature Importance}).
The Causal Feature Importance ($CFI^y_{x_j}$) measures the influence of the change of feature $x_j$ on the outcome $Y$, which can be defined as $CFI^y_{x_j} = \E_{x_j} \Bigr[ \E [Y(x_j)] - \min_{x_j} \E[Y(x_j)] \Bigr] $.
\label{def: causal importance}
\end{definition}

Due to the absence of any prior information, we assume that the $x_j$ is equally likely to be perturbed to any value between $[{low}^{x_j}, {high}^{x_j}]$, i.e., $x_j \thicksim U({low}^{x_j}, {high}^{x_j})$, where $[{low}^{x_j}, {high}^{x_j}]$ is the domain of $x_j $. We use the discrete/continuous uniform distribution, which represents the maximum entropy distribution among all distributions in a given interval. If more information about the distribution of interventions performed by the “external” doer is known, this could be incorporated instead of a uniform distribution.

Due to the fluctuating response curve over the entire range of feature values, the local variation of causal interventional expectation is more crucial for the model explanation. 

\theoremstyle{definition}
\begin{definition}({Locally Causal Positive/Negative/Neutral Gradient}). Locally Causal Gradient is defined as the gradient of the causal response curve, which can be positive, negative, or neutral. 
\label{def: causal attribution}
\end{definition}

\theoremstyle{definition}
\begin{definition}({Causal Feature Attribution}).
The Causal Feature Attribution ($CFA^y_{x_j}$) measures the causal attribution of input feature $x_j$ for output $y$, which can be defined as $ CFA^y_{x_j} =  \E [Y(x_j)] - {baseline}_{x_j}  $.
\label{def: causal attribution}
\end{definition}

${baseline}_{x_j}$ has three forms: $(1)$ the decision boundary of the neural network, where predictions are neutral, such as a probability with value $0.5$ in a binary classification task; $(2)$ a specific value of the feature, i.e., $\Tilde{x}_j$ that the expert assigned according to domain knowledge, such as $ CFA^y_{x_j} =  \E [Y(x_j)] - \E [Y(\Tilde{x}_j)]$; $(3)$ the average $\E [Y(x_j)]$ over the values of $x_j$ as the baseline value for, i.e., $ CFA^y_{x_j} =  \E [Y(x_j)] - \E_{x_j} \Bigr[ \E [Y(x_j)] \Bigr]$. The third form is also defined as Causal Attribution in \cite{chattopadhyay2019neural}. However, the ``positively causal'' ($CFA \geq 0$) is only the relative difference between the average $\E [Y(x_j)]$ over the values of $x_j$, which does not have any practical significance on the contribution of $x_j$ to $y$. In the binary risk classification task, we set ${baseline}_{x_j}=0$ so that causal feature attribution is the same as the causal interventional expectation, which can measure the probability of risk between 0 (negative) and 1 (positive).

\subsection{Risk Prediction}

For now, the original values of each feature can be replaced by the corresponding causal feature attribution that represents the amount of contribution the value of this feature can make to the risk prediction outcome. Thus, the original data $(X,Y)$ containing multifaceted information has been transformed into data $(\mu(X),Y)$ with causal feature attribution on a common scale for this specific risk prediction. Then, a general neural network can be used to produce a trustworthy outcome in line with the causal explanations.

\section{Theoretical Analysis}

In the task-driven causal feature distillation, the original feature values are replaced by the causal expectation estimation for each feature. In this section, we provide a comprehensive theoretical analysis of the unbiased causal expectation estimation based on the propensity score method \cite{chu2023estimating}.

We define the set of risk minimizers as
$\mc{H}^*_{\alpha}:= \{\alpha: K(\alpha) = K(\alpha^*)\}$, where the $K$ is the risk function of propensity score estimation with adaptive group LASSO. The parameters in the first hidden layer for $g_\alpha(X_{i,-j})$ are divided into two groups $ m =  \alpha_\mathcal{I} \bigcup  \alpha_\mathcal{S}$ and $ e=  \alpha_\mathcal{C} \bigcup  \alpha_\mathcal{P}$, which correspond to $v = \beta_\mathcal{I} \bigcup \beta_\mathcal{S}$ and $u = \beta_\mathcal{C} \bigcup \beta_\mathcal{P}$. The following Theorem proves the consistency of estimator and variable selection under adaptive group LASSO:
\begin{theorem}
Let $\gamma>0$, $\epsilon>0$, $\lambda_n = \mathcal{O}(n^{-1/4})$, and $\theta_n =\Omega (n^{-\gamma/(4\nu -4)+ \epsilon})$, for any $\delta >0$. Then there exists $N_\delta$ such that for $n > N_\delta$, $d(\hat \alpha_n, \mc{H^*_{\alpha}})  \le C \left( \frac{\log n}{n}\right)^{\frac{1}{4(\nu-1)}} \quad \text{and} \quad 
\|\hat{m}_{\hat \alpha_n}\| = 0$ with probability at least $1 -\delta$, where $\hat{m}_{\hat \alpha_n} = \hat{\alpha}_\mathcal{I} \bigcup \hat{\alpha}_\mathcal{S}$.
\label{thm:main}
\end{theorem}

Then, the following Theorem proves that the causal expectation estimation based on the propensity score is unbiased.

\begin{theorem}
\label{thm:bigtheorem}
(Bias Removal with Propensity Score) Suppose that
assignment to the intervention $X_j$ is weakly unconfounded given background variables $X_{-j}$. Then \\
(i) $\sigma(x_j, e) = \E[Y(x_j)|e(x_j, X_{-j}) = e] = \E[Y|X_j = x_j,E = e]$. \\
(ii) $\mu(x_j) = \E[Y(x_j)] = \E[\sigma(x_j, e(x_j, X_{-j})]$.
\end{theorem}

\section{Experiments}

In this section, we conduct experiments on synthetic and real datasets, including causal effect estimation benchmarks and synthetic and real datasets for risk prediction tasks to evaluate the following aspects: (1) Our proposed method based on relational graph construction and adaptive group Lasso PS estimation can ensure the accuracy of causal feature attribution estimation; (2) The precision and recall of the risk prediction task are significantly improved. 

\subsection{Causal Effect Estimation Experiments}
We conduct the causal effect estimation experiments on the News dataset with different interventions and compare our TDCFD model with eleven baselines.

\begin{table*}[t]
  \centering
  \scalebox{0.9}{
  \begin{tabular}{lllll}
    \toprule
    \multicolumn{1}{c}{}  & \multicolumn{1}{c}{News-2}  & \multicolumn{1}{c}{News-4} & \multicolumn{1}{c}{News-8}   & \multicolumn{1}{c}{News-16}                    \\
    \multicolumn{1}{c}{Method}  & \multicolumn{1}{c}{$\epsilon_\text{ATE}$}  & \multicolumn{1}{c}{$\epsilon_\text{mATE}$} & \multicolumn{1}{c}{$\epsilon_\text{mATE}$}   & \multicolumn{1}{c}{$\epsilon_\text{mATE}$ }                    \\
    \midrule
    kNN      & $7.83\pm2.55$  & $19.40 \pm3.12$  & $15.11\pm2.34$   &  $17.27\pm2.10$  \\
    PSM       & $4.89\pm 2.39$  & $30.19\pm2.47$  & $22.09\pm1.98$ & $18.81\pm1.74$ \\
    RF     & $5.50\pm 1.20$   & $18.03\pm3.18$ & $12.40\pm2.29$   & $15.91\pm2.00$  \\
    CF     & $4.02\pm 1.33$  & $13.54\pm 2.48$ & $9.70\pm1.91$   & $8.37\pm 1.76$  \\
    BART       & $5.40\pm1.53$   & $17.14\pm 3.51$ & $14.80\pm2.56$  & $17.50\pm2.49$\\
    GANITE      & $4.65\pm 2.12$    & $13.84\pm2.69$ & $11.20\pm 2.84$    & $13.20\pm3.28$  \\
    PD      & $4.69\pm3.17$   & $8.47\pm4.51$   & $7.29\pm2.97$   & $10.65\pm2.22$ \\
    TARNET    & $4.58\pm1.29$   & $13.63\pm2.18$  & $9.38\pm 1.92$    & $8.30\pm1.66$  \\
    CFRNET     & $ 4.54\pm1.48$    & $12.96\pm 1.69$   & $8.79\pm 1.68$     & $8.05\pm1.40$   \\
    SITE    & $4.53\pm1.32$  & $12.75\pm1.88$   & $9.01\pm1.86$    &  $8.63\pm1.41$ \\
    PM   & $\textbf{3.99} \pm \textbf{1.01}$    & $10.04 \pm 2.71$  & $6.51\pm 1.66$  & $ 5.76\pm1.33$ \\
    \midrule
    TDCFD  & $4.25\pm0.98$    & $\textbf{8.77}\pm\textbf{2.49}$   & $\textbf{5.93}\pm\textbf{1.25}$ &   $\textbf{5.04}\pm\textbf{1.19}$\\
   
    \bottomrule
  \end{tabular}}
  \caption{Performance on News data sets. We present the mean $\pm$ standard deviation for $\epsilon_\text{ATE}$ and $\epsilon_\text{mATE}$ on the test sets. We list the available results reported by the original authors \cite{schwab2018perfect}. }
  \label{Newsate}
\end{table*}

\begin{table*}[t]
  \centering
  \scalebox{0.9}{
  \begin{tabular}{lllllll}
    \toprule
    \multicolumn{1}{c}{}  & \multicolumn{3}{c}{ Synthetic data}  & \multicolumn{3}{c}{Real corporate risk data} \\
    \cmidrule(lr){2-4}  \cmidrule(lr){5-7}   

    Method & Accuracy & Precision  & Recall & 
    Accuracy & Precision  & Recall \\
    \midrule
    LR      & $0.92$  & $0.64 $  & $0.54$   &  $0.83$ & $0.21$ & $0.16$\\
    SVM       & $0.94$  & $0.68$  & $0.65$ & $0.87$ & $0.40$ & $0.27$ \\
    KNN     & $0.91$   & $0.55$ & $0.60$   & $0.91$   & $0.62$ & $0.47$\\
    RF     & $0.95$  & $0.72$ & $0.78$   & $0.90$   & $0.60$ & $0.43$\\
    XGBoost       & $0.94$   & $0.67$ & $0.83$  & $0.91$ & $0.61$ & $0.63$\\
    DNN       & $0.95$    & $0.73$ & $0.80$    & $0.93$   & $0.70$ & $0.66$\\
    Transformer     & $0.96$   & $0.77$   & $0.85$   & $0.93$ & $0.71$ & $0.71$\\

    \midrule
    $\textbf{TDCFD}$  & $\textbf{0.97}$    & $\textbf{0.82}$   & $\textbf{0.90}$ &   $\textbf{0.96}$ &   $\textbf{0.86}$ &   $\textbf{0.80}$\\
    
    \bottomrule
  \end{tabular}}
  \caption{Performance on synthetic risk prediction task and real corporate risk prediction task.}
  \label{main result}
\end{table*}

\begin{table*}[t!]
  \centering
  \scalebox{0.9}{
  \begin{tabular}{lllllll}
    \toprule
    \multicolumn{1}{c}{}  & \multicolumn{3}{c}{No hidden variable}  & \multicolumn{3}{c}{Hidden variable} \\
    \cmidrule(lr){2-4}  \cmidrule(lr){5-7}   

    Method & Accuracy & Precision  & Recall & 
    Accuracy & Precision  & Recall \\
    \midrule
 
    DNN       & $0.98$   & $0.89$ & $0.92$  & $0.94$& $0.72\downarrow$& $0.65\downarrow$\\
    XGBoost      & $\textbf{0.99}$   & $\textbf{0.91}$   & $\textbf{0.96}$   & $0.95$ & $0.78\downarrow$& $0.70\downarrow$\\

    \midrule
    $\textbf{TDCFD}$  & $0.98$    & $0.90$   & $0.95$ &   $\textbf{0.97}$&   $\textbf{0.82}$&   $\textbf{0.89}$\\
    
    \bottomrule
  \end{tabular}}
  
  \caption{Performance on synthetic risk prediction data with and without hidden variables.}
  \label{Ablation result}
\end{table*}

\begin{figure}[h]
  \centering
  \includegraphics[width=0.5\linewidth]{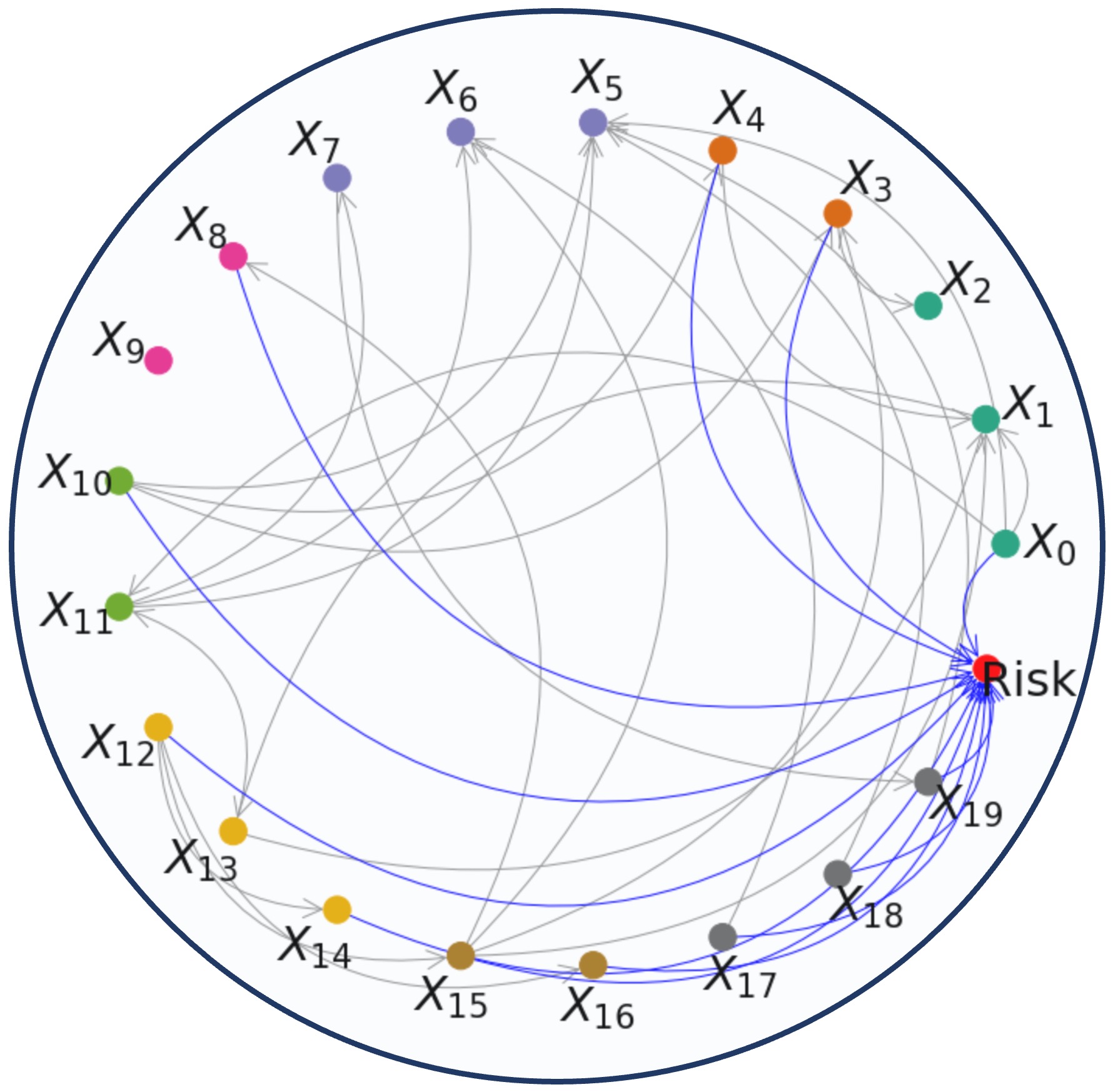}
 
  \caption{The directed acyclic graph of risk data generation.}
 
  \label{fig:generation}
\end{figure}

\begin{figure}
  \centering
  \includegraphics[width=0.95\linewidth]{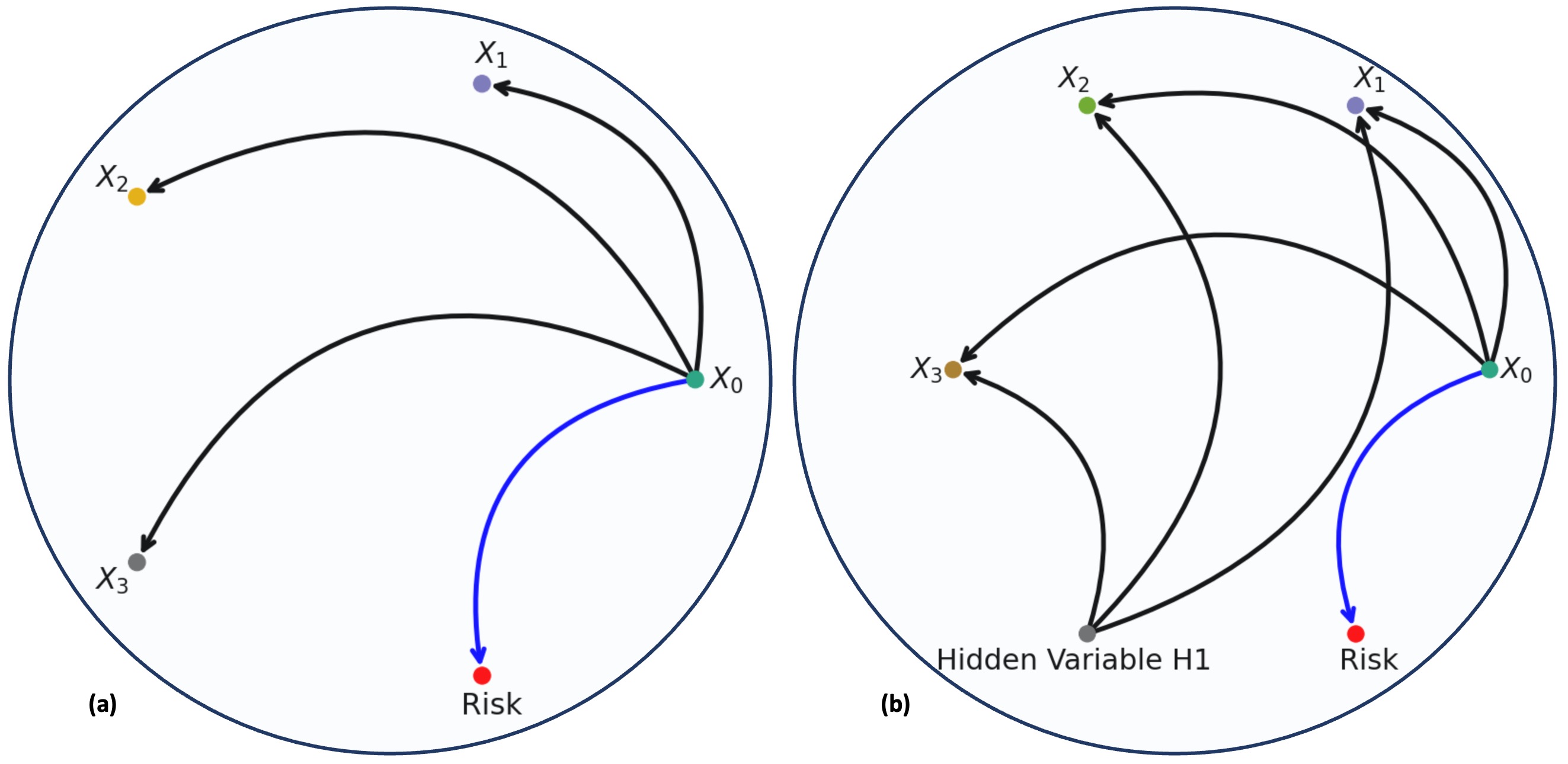}
 
  \caption{The DAGs of risk data generation with and without hidden variables.}
 
  \label{fig:add hidden}
\end{figure}

\noindent\textbf{Results.} We adopt the commonly used evaluation metric, i.e., the error in average treatment effect (ATE) estimation defined as $\epsilon_\text{ATE}  = |\text{ATE} - \widehat{\text{ATE}}|$, where $\widehat{\text{ATE}}$ is an estimated \text{ATE}. In addition, for the evaluation of multiple interventions, we follow the definitions in ~\cite{schwab2018perfect}, where $\epsilon_\text{ATE}$ can be extended to multiple interventions by averaging ATE between every possible pair of interventions. It is defined as $\epsilon_\text{mATE}  = \frac{1}{\binom {k}2}\sum_{i=0}^{k-1}\sum_{j=0}^{i-1}\epsilon_{\text{ATE},i,j}$, where $k$ is the number of intervention options.

Table~\ref{Newsate} shows the performance of our method and baseline methods on the News datasets with $2, 4, 8, \text{and} \ 16$ intervention options. TDCFD achieves the best performance with respect to $\epsilon_\text{ATE}$ on News datasets with $4, 8, \text{and} \ 16$ intervention options. The results of these benchmarks for causal effects estimation can demonstrate that our method is capable of precisely estimating causal effects.

\subsection{Experiments of Risk Prediction on Synthetic}

\noindent\textbf{Simulation Procedure.} 
Because, in the real observational data, the true data generation procedure is unknown, we cannot effectively evaluate the explainability and the true feature contributions.

We generate a synthetic dataset that can not only reflect the complexity of real data but also help to explore the reason why our model can outperform the general machine learning models for the risk prediction task. As shown in Fig. \ref{fig:generation}, our synthetic data includes $20$ features and a binary risk label. In order to incorporate the underlying causal relationships among the features and between the features and the risk outcome, we randomly generate a directed acyclic graph (DAG) to represent the conditional dependency relationships and then utilize the Bayesian networks \cite{heckerman2008tutorial} to simulate the data. Bayesian networks are a type of probabilistic graphical model that uses Bayesian inference for probability computations. It aims to model conditional dependence, and therefore causation, by representing conditional dependence by edges in a directed graph. Data is simulated from a Bayes net by first sampling from each of the root nodes, then followed by the children conditional on their parents until data for all nodes have been drawn. To realistically simulate the risk data, we generate $1,000$ samples with the positive label and $9,000$ with the negative label.

\noindent\textbf{Baseline Methods.} We apply some classical classification models to this risk prediction task, such as Logistic Regression (LR), Support Vector Machine (SVM) \cite{suykens1999least}, K-Nearest Neighbours (KNN) \cite{cunningham2021k}, Random Forest (RF) \cite{breiman2001random}, DNN \cite{larochelle2009exploring}, Transformer \cite{vaswani2017attention}, and XGBoost \cite{chen2016xgboost}.

\noindent\textbf{Evaluation Metrics.}
To evaluate the effectiveness of a model, we adopt Precision ($\frac{TP}{TP+FP}$), Recall ($\frac{TP}{TP+FN}$), and Accuracy ($\frac{TP+TN}{Total}$). Both precision and recall are defined in terms of the positive class. Precision measures the quality of model predictions for positive class and recall, on the other hand, measures how well the model did for the actual observations of the positive class. Compared to accuracy, precision and recall are more important in the risk prediction task.

\noindent\textbf{Results.} 
Table~\ref{main result} shows that TDCFD achieves the best performance with respect to precision and recall in the synthetic data experiment. To further explore the reason why there exist large differences in precision and recall between our model and the baseline models, we performed ablation studies on two more datasets by predicting the risk outcome based on four observed variables ($X_0$, $X_1$, $X_2$, and $X_3$). The first one (Fig.\ref{fig:add hidden} (a)) contains $4$ feature variables and a risk outcome variable, where only $X_0$ is the cause of the outcome also related to $X_1$, $X_2$, and $X_3$. In the second data (Fig.\ref{fig:add hidden} (b)), except for the four observed feature variables, there exists another hidden variable $H1$. $X_1$, $X_2$, and $X_3$ all depend on this hidden variable $H1$. According to Table \ref{Ablation result}, we can find that based on the spurious correlations in the first data, the positive samples can still be accurately captured, but in the second data, the precision and recall decrease dramatically due to ignorance of the real cause.

\subsection{Experiments of Risk Prediction on Real Data}

\noindent\textbf{Real Data.} 
To evaluate the model performance for risk prediction tasks, we adopt a real dataset collected from Alipay, the top Fintech company that offers billions of customers equal access to sustainable financial services and capital. This corporate risk data includes $16,409$ observations with $1,867$ positive samples and $14,542$ negative samples. It contains $114$ feature variables, such as corporate financial statement data, public opinion data, corporate event data, and so on. The baseline methods and evaluation metrics are identical to those in synthetic data experiments.

\begin{figure}
  \centering
  \includegraphics[width=0.7\linewidth]{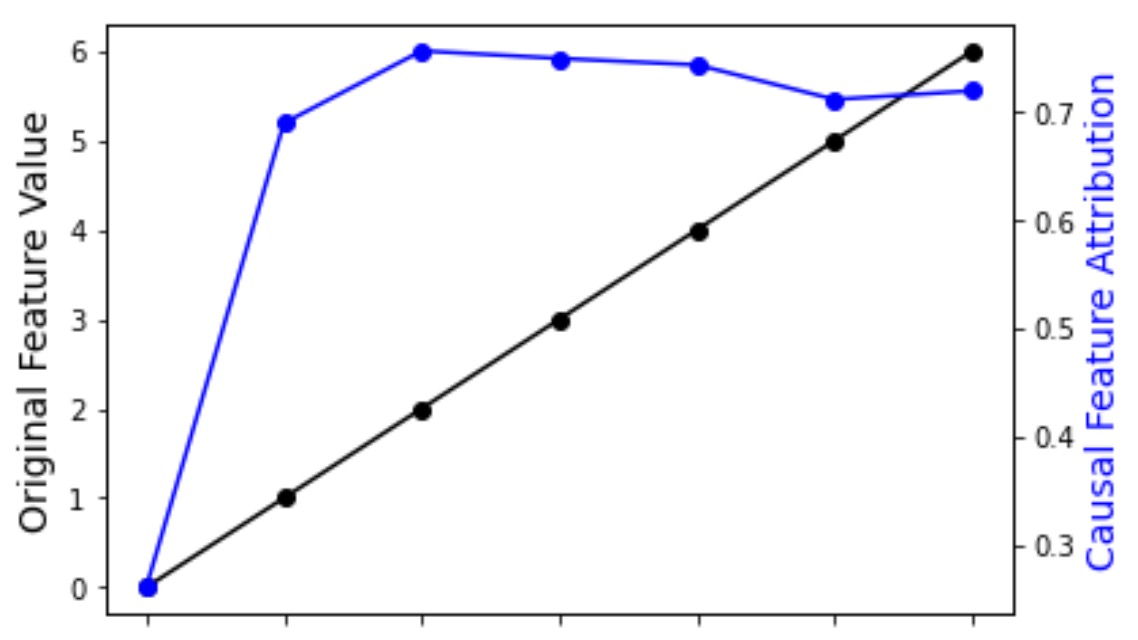}
 
  \caption{Original values and causal feature attributions.}
 
  \label{fig:compare}
\end{figure}

\noindent\textbf{Results.} Table~\ref{main result} shows the performance of our method and baseline methods on the real corporate risk prediction task. TDCFD achieves the best performance with respect to precision and recall. To figure out the reasons for the model's performance, we did two studies on the original feature data (original feature values) and causal feature distilled data, where original feature values are replaced by causal feature attributions. We exhibit a typical categorical feature with original values and causal feature attributions in Fig. \ref{fig:compare}. 
We can find the original values are uniformly increased from $0$ to $6$, but there is a huge gap between the first value and other $5$ values in this feature's causal feature attribution range. The causal feature attributions for the original values from $1$ to $6$ are very close and have similar risk probabilities. However, the original data cannot reflect such information. In addition, we did the t-test for continuous variables and the chi-square test for categorical variables for both original feature data and causal feature distilled data. In the original feature data, 64 variables significantly differ between positive and negative classes. However, in the causal feature distilled data, there are only 52 variables that are significantly different. Furthermore, the 52 variables do not all come from the 64 variables of the original feature data. Therefore, the TDCFD filters out a part of spurious correlations and discovers some new causal relationships that did not appear in the original data.

\section{Conclusion}

We propose a novel Task-Driven Causal Feature Distillation model (TDCFD) for trustworthy risk predictions, which incorporates the POF to distill causal feature contributions and make predictions based on them. We conduct comprehensive experiments on both synthetic and real datasets to illustrate our model can perform well in risk prediction tasks with significantly improved precision and recall and generate causal-based interpretability.

\bibliography{aaai24}

\begin{thebibliography}{38}
\providecommand{\natexlab}[1]{#1}

\bibitem[{Bishop(1994)}]{bishop1994mixture}
Bishop, C.~M. 1994.
\newblock Mixture density networks.

\bibitem[{Breiman(2001)}]{breiman2001random}
Breiman, L. 2001.
\newblock Random forests.
\newblock \emph{Machine learning}, 45(1): 5--32.

\bibitem[{Chakraborty et~al.(2018)Chakraborty, Alam, Dey, Chattopadhyay, and Mukhopadhyay}]{chakraborty2018adversarial}
Chakraborty, A.; Alam, M.; Dey, V.; Chattopadhyay, A.; and Mukhopadhyay, D. 2018.
\newblock Adversarial attacks and defences: A survey.
\newblock \emph{arXiv preprint arXiv:1810.00069}.

\bibitem[{Chattopadhyay et~al.(2019)Chattopadhyay, Manupriya, Sarkar, and Balasubramanian}]{chattopadhyay2019neural}
Chattopadhyay, A.; Manupriya, P.; Sarkar, A.; and Balasubramanian, V.~N. 2019.
\newblock Neural network attributions: A causal perspective.
\newblock In \emph{International Conference on Machine Learning}, 981--990. PMLR.

\bibitem[{Chen and Guestrin(2016)}]{chen2016xgboost}
Chen, T.; and Guestrin, C. 2016.
\newblock Xgboost: A scalable tree boosting system.
\newblock In \emph{Proceedings of the 22nd acm sigkdd international conference on knowledge discovery and data mining}, 785--794.

\bibitem[{Chen, Rezayi, and Li(2023)}]{chen2023more}
Chen, Z.; Rezayi, S.; and Li, S. 2023.
\newblock More Knowledge, Less Bias: Unbiasing Scene Graph Generation with Explicit Ontological Adjustment.
\newblock In \emph{Proceedings of the IEEE/CVF Winter Conference on Applications of Computer Vision}, 4023--4032.

\bibitem[{Chu et~al.(2023{\natexlab{a}})Chu, Claridy, Cordero, Li, and Rathbun}]{chu2023estimating}
Chu, Z.; Claridy, M.; Cordero, J.; Li, S.; and Rathbun, S.~L. 2023{\natexlab{a}}.
\newblock Estimating propensity scores with deep adaptive variable selection.
\newblock In \emph{Proceedings of the 2023 SIAM International Conference on Data Mining (SDM)}, 730--738. SIAM.

\bibitem[{Chu et~al.(2023{\natexlab{b}})Chu, Guo, Zhou, Wang, Yu, Chen, Xu, Lu, Cui, Li et~al.}]{chu2023data}
Chu, Z.; Guo, H.; Zhou, X.; Wang, Y.; Yu, F.; Chen, H.; Xu, W.; Lu, X.; Cui, Q.; Li, L.; et~al. 2023{\natexlab{b}}.
\newblock Data-Centric Financial Large Language Models.
\newblock \emph{arXiv preprint arXiv:2310.17784}.

\bibitem[{Chu et~al.(2023{\natexlab{c}})Chu, Huang, Li, Chu, and Li}]{chu2023causal}
Chu, Z.; Huang, J.; Li, R.; Chu, W.; and Li, S. 2023{\natexlab{c}}.
\newblock Causal effect estimation: Recent advances, challenges, and opportunities.
\newblock \emph{arXiv preprint arXiv:2302.00848}.

\bibitem[{Chu, Rathbun, and Li(2020)}]{chu2020matching}
Chu, Z.; Rathbun, S.~L.; and Li, S. 2020.
\newblock Matching in selective and balanced representation space for treatment effects estimation.
\newblock In \emph{Proceedings of the 29th ACM International Conference on Information \& Knowledge Management}, 205--214.

\bibitem[{Chu, Rathbun, and Li(2021)}]{chu2021graph}
Chu, Z.; Rathbun, S.~L.; and Li, S. 2021.
\newblock Graph infomax adversarial learning for treatment effect estimation with networked observational data.
\newblock In \emph{Proceedings of the 27th ACM SIGKDD Conference on Knowledge Discovery \& Data Mining}, 176--184.

\bibitem[{Chu, Rathbun, and Li(2022)}]{chu2022learning}
Chu, Z.; Rathbun, S.~L.; and Li, S. 2022.
\newblock Learning infomax and domain-independent representations for causal effect inference with real-world data.
\newblock In \emph{Proceedings of the 2022 SIAM International Conference on Data Mining (SDM)}, 433--441. SIAM.

\bibitem[{Cunningham and Delany(2021)}]{cunningham2021k}
Cunningham, P.; and Delany, S.~J. 2021.
\newblock K-nearest neighbour classifiers-a tutorial.
\newblock \emph{ACM Computing Surveys (CSUR)}, 54(6): 1--25.

\bibitem[{Greenland(2008)}]{greenland2008invited}
Greenland, S. 2008.
\newblock Invited commentary: variable selection versus shrinkage in the control of multiple confounders.
\newblock \emph{American journal of epidemiology}, 167(5): 523--529.

\bibitem[{Guan et~al.(2023{\natexlab{a}})Guan, Hu, Zhou, Zhang, Li, and Liu}]{guan2023badsam}
Guan, Z.; Hu, M.; Zhou, Z.; Zhang, J.; Li, S.; and Liu, N. 2023{\natexlab{a}}.
\newblock Badsam: Exploring security vulnerabilities of sam via backdoor attacks.
\newblock \emph{arXiv preprint arXiv:2305.03289}.

\bibitem[{Guan et~al.(2023{\natexlab{b}})Guan, Sun, Du, and Liu}]{Guan}
Guan, Z.; Sun, L.; Du, M.; and Liu, N. 2023{\natexlab{b}}.
\newblock Attacking Neural Networks with Neural Networks: Towards Deep Synchronization for Backdoor Attacks.
\newblock In \emph{Proceedings of the 32nd ACM International Conference on Information and Knowledge Management}, CIKM '23, 608–618. New York, NY, USA: Association for Computing Machinery.
\newblock ISBN 9798400701245.

\bibitem[{Heckerman(2008)}]{heckerman2008tutorial}
Heckerman, D. 2008.
\newblock A tutorial on learning with Bayesian networks.
\newblock \emph{Innovations in Bayesian networks}, 33--82.

\bibitem[{Larochelle et~al.(2009)Larochelle, Bengio, Louradour, and Lamblin}]{larochelle2009exploring}
Larochelle, H.; Bengio, Y.; Louradour, J.; and Lamblin, P. 2009.
\newblock Exploring strategies for training deep neural networks.
\newblock \emph{Journal of machine learning research}, 10(1).

\bibitem[{Li and Chu(2023)}]{li2023machine}
Li, S.; and Chu, Z. 2023.
\newblock \emph{Machine Learning for Causal Inference}.
\newblock Springer Nature.

\bibitem[{Li et~al.(2022)Li, Jiang, Li, and Xia}]{li2022backdoor}
Li, Y.; Jiang, Y.; Li, Z.; and Xia, S.-T. 2022.
\newblock Backdoor learning: A survey.
\newblock \emph{IEEE Transactions on Neural Networks and Learning Systems}.

\bibitem[{Lin, Feng, and Li(2015)}]{lin2015regularization}
Lin, W.; Feng, R.; and Li, H. 2015.
\newblock Regularization methods for high-dimensional instrumental variables regression with an application to genetical genomics.
\newblock \emph{Journal of the American Statistical Association}, 110(509): 270--288.

\bibitem[{Liu et~al.(2023)Liu, Wu, Hu, Zhao, Zhao, Zhang, Dai, Chen, Shen, Li et~al.}]{liu2023pharmacygpt}
Liu, Z.; Wu, Z.; Hu, M.; Zhao, B.; Zhao, L.; Zhang, T.; Dai, H.; Chen, X.; Shen, Y.; Li, S.; et~al. 2023.
\newblock Pharmacygpt: The ai pharmacist.
\newblock \emph{arXiv preprint arXiv:2307.10432}.

\bibitem[{Luan and Tsai(2021)}]{luan2021review}
Luan, H.; and Tsai, C.-C. 2021.
\newblock A review of using machine learning approaches for precision education.
\newblock \emph{Educational Technology \& Society}, 24(1): 250--266.

\bibitem[{Myers et~al.(2011)Myers, Rassen, Gagne, Huybrechts, Schneeweiss, Rothman, Joffe, and Glynn}]{myers2011effects_instrumental}
Myers, J.~A.; Rassen, J.~A.; Gagne, J.~J.; Huybrechts, K.~F.; Schneeweiss, S.; Rothman, K.~J.; Joffe, M.~M.; and Glynn, R.~J. 2011.
\newblock Effects of adjusting for instrumental variables on bias and precision of effect estimates.
\newblock \emph{American journal of epidemiology}, 174(11): 1213--1222.

\bibitem[{Rosenbaum and Rubin(1983)}]{rosenbaum1983central}
Rosenbaum, P.~R.; and Rubin, D.~B. 1983.
\newblock The central role of the propensity score in observational studies for causal effects.
\newblock \emph{Biometrika}, 70(1): 41--55.

\bibitem[{Rubin(1974)}]{rubin1974estimating}
Rubin, D.~B. 1974.
\newblock Estimating causal effects of treatments in randomized and nonrandomized studies.
\newblock \emph{Journal of educational Psychology}, 66(5): 688.

\bibitem[{Sauer et~al.(2013)Sauer, Brookhart, Roy, and VanderWeele}]{sauer2013review}
Sauer, B.~C.; Brookhart, M.~A.; Roy, J.; and VanderWeele, T. 2013.
\newblock A review of covariate selection for non-experimental comparative effectiveness research.
\newblock \emph{Pharmacoepidemiology and drug safety}, 22(11): 1139--1145.

\bibitem[{Schisterman, Cole, and Platt(2009)}]{schisterman2009overadjustment}
Schisterman, E.~F.; Cole, S.~R.; and Platt, R.~W. 2009.
\newblock Overadjustment bias and unnecessary adjustment in epidemiologic studies.
\newblock \emph{Epidemiology (Cambridge, Mass.)}, 20(4): 488.

\bibitem[{Schwab, Linhardt, and Karlen(2018)}]{schwab2018perfect}
Schwab, P.; Linhardt, L.; and Karlen, W. 2018.
\newblock Perfect match: A simple method for learning representations for counterfactual inference with neural networks.
\newblock \emph{arXiv preprint arXiv:1810.00656}.

\bibitem[{Shi et~al.(2023)Shi, Zhou, Letcher, Hitt, Kanno, Futamura, Kishida, Morita, and Li}]{shi2023aging}
Shi, W.; Zhou, Z.; Letcher, B.~H.; Hitt, N.; Kanno, Y.; Futamura, R.; Kishida, O.; Morita, K.; and Li, S. 2023.
\newblock Aging Contrast: A Contrastive Learning Framework for Fish Re-identification Across Seasons and Years.
\newblock In \emph{Australasian Joint Conference on Artificial Intelligence}, 252--264. Springer.

\bibitem[{Shortreed and Ertefaie(2017)}]{shortreed2017outcome}
Shortreed, S.~M.; and Ertefaie, A. 2017.
\newblock Outcome-adaptive lasso: Variable selection for causal inference.
\newblock \emph{Biometrics}, 73(4): 1111--1122.

\bibitem[{Suykens and Vandewalle(1999)}]{suykens1999least}
Suykens, J.~A.; and Vandewalle, J. 1999.
\newblock Least squares support vector machine classifiers.
\newblock \emph{Neural processing letters}, 9(3): 293--300.

\bibitem[{Vaswani et~al.(2017)Vaswani, Shazeer, Parmar, Uszkoreit, Jones, Gomez, Kaiser, and Polosukhin}]{vaswani2017attention}
Vaswani, A.; Shazeer, N.; Parmar, N.; Uszkoreit, J.; Jones, L.; Gomez, A.~N.; Kaiser, {\L}.; and Polosukhin, I. 2017.
\newblock Attention is all you need.
\newblock \emph{Advances in neural information processing systems}, 30.

\bibitem[{Wang et~al.(2023)Wang, Guo, Li, and Fu}]{wang2023towards}
Wang, Y.; Guo, D.; Li, S.; and Fu, Y. 2023.
\newblock Towards Explainable Visual Anomaly Detection.
\newblock \emph{arXiv preprint arXiv:2302.06670}.

\bibitem[{Wilson and Reich(2014)}]{wilson2014confounder}
Wilson, A.; and Reich, B.~J. 2014.
\newblock Confounder selection via penalized credible regions.
\newblock \emph{Biometrics}, 70(4): 852--861.

\bibitem[{Yao et~al.(2021)Yao, Chu, Li, Li, Gao, and Zhang}]{yao2021survey}
Yao, L.; Chu, Z.; Li, S.; Li, Y.; Gao, J.; and Zhang, A. 2021.
\newblock A survey on causal inference.
\newblock \emph{ACM Transactions on Knowledge Discovery from Data (TKDD)}, 15(5): 1--46.

\bibitem[{Zhu et~al.(2023)Zhu, Guo, Qi, Chu, Yu, and Li}]{zhu2023trustworthy}
Zhu, R.; Guo, D.; Qi, D.; Chu, Z.; Yu, X.; and Li, S. 2023.
\newblock Trustworthy Representation Learning Across Domains.
\newblock \emph{arXiv preprint arXiv:2308.12315}.

\bibitem[{Zou(2006)}]{zou2006adaptive}
Zou, H. 2006.
\newblock The adaptive lasso and its oracle properties.
\newblock \emph{Journal of the American statistical association}, 101(476): 1418--1429.

\end{thebibliography}

\end{document}